\documentclass{article}
\usepackage{spconf,amsmath,epsfig}
\usepackage{amsmath,amssymb}
\usepackage{dcolumn}
\usepackage{hyperref}
\usepackage{graphicx}
\usepackage{float}
\usepackage{subcaption}
\usepackage[T1]{fontenc}
\usepackage[colorinlistoftodos]{todonotes}
\usepackage{listings}
\usepackage{blindtext}
\usepackage{relsize}
\usepackage{ragged2e} 
\usepackage{animate}
\usepackage{booktabs}
\usepackage{multirow}

\makeatother

\let\OLDthebibliography\thebibliography
\renewcommand\thebibliography[1]{
  \OLDthebibliography{#1}
  \setlength{\parskip}{0pt}
  \setlength{\itemsep}{0pt plus 0.3ex}
}

\begin{document}\sloppy

% Example definitions.
% --------------------
\def\x{{\mathbf x}}
\def\L{{\cal L}}

% Title.
% ------
\title{Learning Long Term Style Preserving Blind Video Temporal Consistency}
%
% Single address.
% ---------------
\name{Hugo Thimonier*, Julien Despois*, Robin Kips*$^{,\dagger}$, Matthieu Perrot*}
%Address and e-mail should NOT be added in the submission paper. They should be present only in the camera ready paper. 
\address{* L'Or\'eal Research and Innovation, France \\ 
        $\dagger$ LTCI, T\'el\'ecom Paris, Institut Polytechnique de Paris, France \\
        \{firstname.lastname\}@rd.loreal.com}

\maketitle

\begin{abstract}
When trying to independently apply image-trained algorithms to successive frames in videos, noxious flickering tends to appear. State-of-the-art post-processing techniques that aim at fostering temporal consistency, generate other temporal artifacts and visually alter the style of videos. We propose a post-processing model, agnostic to the transformation applied to videos (\textit{e.g.} style transfer, image manipulation using GANs, etc.), in the form of a recurrent neural network. Our model is trained using a Ping Pong procedure and its corresponding loss, recently introduced for GAN video generation, as well as a novel style preserving perceptual loss. The former improves long-term temporal consistency learning, while the latter fosters style preservation. We evaluate our model on the DAVIS and videvo.net datasets and show that our approach offers state-of-the-art results concerning flicker removal, and better keeps the overall style of the videos than previous approaches. 
\end{abstract}

\begin{keywords}
Time consistency, deep learning, video post-processing, style preserving perceptual loss
\end{keywords}

\section{Introduction}

Since the first introduction of deep convolutional neural networks (CNN) and  the rise of  GANs, more and more powerful image processing applications have been proposed. A still popular topic in the literature discusses how these image processing techniques can be efficiently extended to videos.

A natural way to transfer image processing algorithms to videos is to perform such transformation frame by frame. However, the latter scheme necessarily causes unpleasant visual artifacts (\textit{e.g.} flickering) originating from temporal inconsistency. For instance, successive frames are not necessarily transformed coherently since image trained algorithms fail to capture the temporal dimension of videos. The literature offers several types of approaches to circumvent such problems: video-to-video translation algorithms and post-processing techniques of per-frame transformed videos. The former consists in conditional video generation algorithms aiming at generating temporally coherent outputs, while the latter post-processes videos transformed using image-trained algorithms to make them temporally coherent. On the one hand, video-to-video translation algorithms exhibit interesting performances even for relatively high-resolution videos, however, they require video datasets which are very scarce especially for supervised algorithms. On the other hand, post-processing techniques offer satisfactory results and do not necessitate any retraining of image processing algorithms, but often fail to reduce localized flickering without generating new visual artifacts and tend to modify the overall color of videos.
\begin{figure}[t!]
\centering
  \centering
   \animategraphics[autoplay,loop,scale=0.5, poster=40]{30}{files/videos/cows_new_folder/frame-}{0}{79}
\caption{Our model successfully removes flickering from the raw processed video without altering the overall style nor the brightness. \cite{lai2018learning} on the contrary seems to remove flickering at the cost of style alteration as the video becomes darker. This video is best seen using Adobe Acrobat Reader.}
  \label{fig:cows}
\end{figure}

We propose a blind post-processing algorithm in the form of a recurrent neural network. The main contributions involved in the model are:
\begin{itemize}
    \item[•] Following up on recent work \cite{park2019arbitrary} in the image style transfer literature, we introduce a novel style preserving perceptual loss which reduces deviation from the overall style of the original processed video. 
    \item[•] We propose to enforce long term consistency by adapting the Ping Pong training procedure \cite{Chu_2020} to this task and its corresponding Ping Pong Loss. 
    \item[•] We empirically highlight some weaknesses of the perceptual distance as a measure of style preservation.
    \item[•] Our model offers state-of-the-art results in flicker removal for any transformation and does not deteriorate the visual quality of videos when enforcing time consistency, nor generate new temporal artifacts.
\end{itemize}
%%%%%%%%%%%%%%%%%%%%%%%%%%%%%%%%%%%%%%%%%%%%%%%%%%%%%%%%%%%%%%
\section{Background and related work}
When independently applying image-trained algorithms (\textit{e.g.} image-trained style transfer algorithms) to successive frames in a video, several temporal artifacts tend to appear. In that context, the recent literature has focused on two approaches to tackle the problem of video processing and temporal inconsistency.

The first approach, namely video-to-video translation, consists in adapting image processing algorithms to take into account the temporal dimension of the input. Most approaches involve computing optical flows during training and consider networks taking several inputs including the current frames (raw and stylized), the previous output frame, and the previous stylized frame warped using the estimated optical flow \cite{Ruder_2016}. Optical flow estimation is computationally costly, thus, later approaches also considered similar methods but circumvent the problem of computing optical flows during the inference stage such as \cite{gao2020}. Recent approaches offer satisfying results but tend to require complex training procedures such as a two-stage progressive training strategy and need to be retrained for each new application considered.

The second method, task-independent post-processing, aims at considering time consistency in a more agnostic manner not focusing on any type of image processing. Blind post-processing consists in (i) first transforming a video using any algorithm and (ii) using an algorithm, independent of the transformation previously performed, to make the video temporally consistent. This approach requires to consider a twofold problem: on the one hand, videos must be made temporally coherent, on the other hand, the style of videos must not be altered.
Recent papers have obtained satisfying results using general approaches totally blind to the filter applied to videos. For instance, \cite{bonneel:hal-01264081} and \cite{yao2017} display very satisfactory results, however, the computational complexity of their proposed algorithm makes it difficult to process long video sequences. 
More recently, in \cite{lai2018learning} the authors propose a neural network to enforce time consistency in videos. The model takes as an input the raw unprocessed video frames, the processed video frames, and the previous output of the recurrent network. The model also makes use of optical flow estimation for temporal consistency loss computation and of the VGG19 \cite{simonyan2014deep} network for perceptual loss computation. Their model offers interesting results but fails to coherently learn long-term temporal consistency and generates new temporal artifacts when the motion of an object changes direction. Moreover, their model tends to alter the overall style of the video once post-processed, thus reducing the quality of the video transformation. Recent work in the style transfer literature has offered possible solutions to foster style preservation when learning temporal coherence.  

Neural style transfer consists in transferring the style of an image (\textit{e.g.} a painting) to another one while preserving the content of the latter image. Style transfer in the form of deep neural networks was first introduced in \cite{Gatys}. Their seminal model makes use of the VGG19 \cite{simonyan2014deep} network to perform style transfer through both a content and style loss. Their approach opened a wide range of applications making use of deep features (\textit{e.g.} VGG19 \cite{simonyan2014deep}) as measures of similarity between images. Recent approaches such as \cite{park2019arbitrary}
have offered a better balance between the structure and the style of the output by using moments of different orders (mean and standard deviation) of the pre-trained features.

\section{Proposed method}
\label{nn}
\subsection{Task}
We argue that post-processing techniques can manage to efficiently remove any temporal artifacts on videos originating from per-frame processing. This approach is all the more appealing since it needs not any retraining for each task; this is a key aspect since it can be challenging to find a satisfying equilibrium in the case of video-to-video translation algorithms. We thus propose a blind post-processing model in the form of a recurrent neural network which improves current state-of-the-art results. Our approach does not require optical flow estimation during inference but necessitates to dispose of the original non-transformed video. 

\subsection{Architecture}
The deep network is comprised of a classical encoder-decoder architecture. The encoder part is composed of two downsampling strided convolutional layers, each followed by Instance Normalization. The encoder is then followed by 5 residual blocks and a ConvLSTM module \cite{NIPS2015_5955}. 
The decoder placed after the ConvLSTM module is composed of two transposed convolutional layers also followed by Instance Normalization. We also include skip-connections via concatenation from the encoder to the decoder.

ConvLSTM modules are well-suited to our problem since they can compress all the previous information into a hidden state which can be used to estimate the current state. This hidden state can capture the spatial information of the whole input sequence and allows the model to learn coherently time consistency. In tandem with temporal losses, ConvLSTM modules were shown to offer very satisfactory results in fostering temporal consistency on video style transfer networks.

Let us denote the original unprocessed frames $\{I_t\}_{t=1,\hdots,T }$, the per-frame processed videos $\{P_t\}_{t=1,\hdots,T }$ and $\{O_t\}_{t=1,\hdots,T }$ the corresponding outputs of the model. Since we consider residual blocks in our network, $O_t$ can be seen as a residual transformation of $P_t$ to make it temporally consistent. To output $O_t$, the network takes as an input $I_t, I_{t-1}, P_t$ and $O_{t-1}$. Formally
\[
O_t = P_t + \mathcal{F}(I_t, I_{t-1}, P_t, O_{t-1})
\]
where $\mathcal{F}(.)$ denotes our network. The recurrent form of the network thus originates from the fact that our network makes use of the previous output to generate the current output. Let us note here that for of each sequence $P_1 = O_1$.

\subsection{Losses}
\label{losse}
The aim of the proposed model is to reduce the temporal inconsistency while maintaining the perceptual similarity in both content and style between the output frame of the model and the processed frame. Therefore, two types of loss must be considered: perceptual losses and temporal losses. 
\subsubsection{Perceptual losses}
%%
%%
%%
%%
%\subsubsection{Content Perceptual Loss}
%\textbf{Content Perceptual Loss}
The pre-trained VGG19 classification network is vastly used in a large number of application as it has been shown to be a good measure of human perception. To preserve the content of the video, as in \cite{lai2018learning} we set the content perceptual loss to be
\begin{equation}
    \label{eq:perceptual_loss}
    \mathcal{L}_p = \sum_{t=2}^T \sum_{i=1}^N \sum_{l} \bigg{|}\bigg{|}\phi_l\big{(}O_t^{(i)}\big{)} - \phi_l\big{(}P_t^{(i)}\big{)} \bigg{|}\bigg{|}_1
\end{equation}
where $O_t^{(i)} \in \mathbb{R}^3$ represent the RGB vector of pixel $i$ in the output frame $O_t$ composed of $N$ pixels, while $\phi_l(.)$ represents the feature activation of the l-th block of the VGG19 network $\phi$ (we consider a sum of all feature activation until the 4th block, \textit{i.e.} \texttt{relu1-2}, \texttt{relu2-2}, \texttt{relu3-3} and \texttt{relu4-3}).

To ensure that our model does not alter the style of the video, we also introduce a novel loss which makes use of the VGG network discussed above. For instance, we consider
\begin{equation}
    \label{eq:perceptual_loss}
    \begin{array}{rcl}
    \mathcal{L}_{sp} &=& \displaystyle{\sum_{t=2}^T \sum_{l}} \{ \big{|}\big{|}\mu\big{(}\phi_l(O_t)\big{)} - \mu\big{(}\phi_l(P_t)\big{)} \big{|}\big{|}_2^2 \\
    & & + \big{|}\big{|}\sigma\big{(}\phi_l(O_t)\big{)} - \sigma\big{(}\phi_l(P_t)\big{)} \big{|}\big{|}_2^2 \}
    \end{array}
\end{equation}
where $\mu(.)$ denotes the average in the channel dimension and $\sigma(.)$ the standard deviation. 
Such approach is well suited to our case since pixel-wise comparisons between $P_t$ and $O_t$ can be too conservative, while considering only moments of the first and second order tend to capture well global style of a frame. 
We also consider its temporal equivalent to avoid excessive deviations from the global style of the raw processed videos. For instance, since images are generated in an online manner, minor deviation from the style of the video tend to accumulate as frames are being processed.
\begin{equation}
    \label{eq:perceptual_loss}
    \begin{array}{lll}
    \mathcal{L}_{sp}^{temp} &=& \displaystyle{\sum_{t=2}^T \sum_{l}} \{ \big{|}\big{|}\mu\big{(} \phi_l(O_{t})\big{)} - \mu\big{(}\phi_l( O_{t-1})\big{)} \big{|}\big{|}_2^2 \\
    & & + \big{|}\big{|}\sigma\big{(}\phi_l( O_{t})\big{)} - \sigma\big{(} \phi_l(O_{t-1})\big{)} \big{|}\big{|}_2^2 \}
    \end{array}
\end{equation}
In both losses, the feature activation considered are \texttt{relu1-2} and \ \texttt{relu2-2}. We thus consider the following style preserving perceptual loss 
\[
\mathcal{L}_{SP} = \mathcal{L}_{sp} + \mathcal{L}_{sp}^{temp} 
\]
\subsubsection{Temporal losses}
During training, we randomly fetch a sequence of $k$ successive frames in a video which we transform to construct a temporal cycle centered around the last frame of the selected sequence. For instance, for each randomly selected sequence in the training dataset $\{P_t, \hdots , P_{t+k-1}, P_{t+k} \} $ the following ping pong sequence is constructed $\{P_t, \hdots , P_{t+k-1}, P_{t+k}, P_{t+k-1}, \hdots, P_t \}$. In the recurrent framework detailed in the previous section, one obtains a similar sequence of output frames
$$
\{O_t, \hdots , O_{t+k-1}, O_{t+k}, O_{t+k-1}', \hdots, O_t' \} 
$$
where frames with <<'>> denote frames constructed in a backward direction. In other words, $O_t'$ is estimated by our network taking as inputs $I_t, I_{t+1}, O_{t+1}'$ and $P_t$.
\noindent Considering such procedure, \cite{Chu_2020} propose the following Ping Pong loss
\begin{equation}
    \label{eq:pingpong}
    \mathcal{L}_{PP} = \sum_{t=1}^{k-1} \big{|}\big{|}O_t - O_t'\big{|}\big{|}_2
\end{equation}
Such loss which combines short term consistency for frames close to $n$, \textit{e.g.} $||O_{n-1} - O_{n-1}'||_2$ and long term consistency for factors far from $n$, \textit{e.g.} $||O_1 - O_1'||_2$, should reduce the accumulation of unwanted features throughout the video.

The short term temporal error is based on the warping error of successive output frames. Notice that the Ping Pong training procedure modifies classical short term temporal loss by involving backward warping error
\begin{equation}
    \label{eq:st_loss}
    \begin{array}{ll}
    \mathcal{L}_{st} = & \displaystyle{\sum_{t=2}^T \sum_{i=1}^N} M_{t,t-1}^{(i)} \big{|}\big{|} O_t^{(i)} - warp(O_{t-1}, F_{t,t-1})^{(i)}  \big{|}\big{|}_1 \\
    & + M_{t,t+1}^{(i)} \big{|}\big{|} O_t'^{(i)} - warp(O_{t+1}', F_{t,t+1})^{(i)}  \big{|}\big{|}_1
    \end{array}
\end{equation}
where $warp(O_{t-1}, F_{t,t-1})$ is the $O_{t-1}$ frame warped to time $t$ using the backward optical flow $F_{t,t-1}$ estimated between $I_t$ and $I_{t-1}$. $M_{t,t-1} = \exp(-\alpha ||I_t - warp(I_{t-1}, F_{t,t-1})||_2^2)$ is the visibility mask calculated from the warping error between the input frame $I_t$ and the warped input frame $warp(I_{t-1}, F_{t,t-1})$ where $\alpha$ is set to be 50 \cite{lai2018learning}. Optical flows are estimated using FlowNet2 \cite{flownet2_ilg017}.

The previously described loss is used to enforce consistency between adjacent frames, however there is no guarantee regarding long-term consistency. The proposed long-term temporal loss consists in the warping error between the first frame of the sequence and all of the output frames of the forward sequence, 
\begin{equation}
    \label{eq:lt_loss}
    \mathcal{L}_{lt} = \sum_{t=2}^T \sum_{i=1}^N M_{t,1}^{(i)} \big{|}\big{|} O_t^{(i)} - warp(O_{1}, F_{t,1})^{(i)}  \big{|}\big{|}_1
\end{equation}

Following up on \cite{chen2020optical} we formulate the hypothesis on stable stylized/processed videos: non-occluded regions should be low-rank representations.  
Consider during training (i) a sequence of $k$ consecutive frames, $\{I_t, \hdots , I_{t+k-1}, I_{t+k} \} $ (ii) a reference time (which we simply set to $k/2$) to which all frames are warped (iii) and occlusion masks at each time step. 
Denoting $R_t$ the Hadamard product (element-wise product) between the occlusion mask and the warped frame, consider
$
\chi = [vec(R_0), \hdots, vec(R_k)]^\top \in \mathbb{R}^{k \times N}
$
where $N = H \times W$ the number of pixels in the image and $vec(.)$ simply denotes the flattening operator of a two-dimensional image. 
Based on the formulated hypothesis, the rank of $\chi$ constructed using the raw input frames $I_t$, $\chi_I$ and using the output frames of the model $O_t$, $\chi_O$ should not be too different from each other. \cite{chen2020optical} propose the low rank loss using the convex relaxation of the rank defined as
\begin{equation}
    \label{eq:lowrank}
    \mathcal{L}_{rank} = (||\chi_I||_* - ||\chi_O||_*)^2
\end{equation}
where $||.||_*$ is the nuclear norm defined as the sum of all singular values of a matrix.
\subsubsection{Overall loss}
 The overall loss is a weighted sum of all previously defined losses
\begin{equation}
    \label{eq:refined_overall_loss}
    \begin{array}{rcl}
    \mathcal{L} & = &\lambda_p\mathcal{L}_{p} + \lambda_{SP}\mathcal{L}_{SP} + \lambda_{st}\mathcal{L}_{st} + \\
    &&  \lambda_{lt} \mathcal{L}_{lt} + \lambda_{rank} \mathcal{L}_{rank} + \lambda_{PP} \mathcal{L}_{PP} 
    \end{array}
\end{equation}
where $\lambda_p$, $\lambda_{SP}$, $\lambda_{st}$, $\lambda_{lt}$, $\lambda_{rank}$ and $\lambda_{PP}$ are hyper-parameters to be defined, that represent the weight given to each loss in the overall training of the network.
%
%%%%%%%%%%%%%%%%%%%%%%%%%%%%%%%%%%%%%%%%%%%%%%%%%%%%%%%%%%%%%%
\section{Experimental results}
\subsection{Evaluation metrics}
\label{pd}
Two features must be evaluated to assess the quality of the proposed model : (i) the output frames $O_t$ must be as close as possible to unprocessed frames $I_t$ in terms of time consistency and (ii) the output frames $O_t$ must visually look like the input processed frames $P_t$. Two types of metrics are used in the literature to evaluate the quality of a model: a time consistency metric based on the warping error and a perceptual similarity metric.

The literature dealing with time consistency has offered several metrics to measure time consistency in a video. The most vastly used is the mean warping error between successive frames of a video. The latter is defined as the difference between a warped frame to time $t$ and the true frame at time $t$. 
Formally, the warping error between two frames $V_t$ and $V_{t+1}$ is defined as
\[
\omega_{t,t+1}^V = \frac{1}{M}\sum_{i=1}^NM_t^{(i)} || V_t^{(i)} - warp(V_{t+1},F_{t,t+1})^{(i)}||_2^2 
\]
where $M_t \in \{0,1\}^{N}$ is the non-occlusion mask indicating non-occluded regions, $N$ is the number of pixels in each frame and $M =  \sum_{i=1}^N M_t^{(i)}$. To estimate those regions, we resort to the method given in \cite{Ruder_2016} which consists in performing a backward-forward consistency check of the optical flows based on the method introduced in \cite{Bro10e}. To obtain a metric for a whole video $V$, one averages the warping error between successive frames over the sequence.

Perceptual similarity is evaluated using the LPIPS metric proposed in \cite{zhang2018unreasonable} to evaluate the perceptual similarity between the raw processed and output videos. 
In \cite{lai2018learning}, the authors resort to the LPIPS metric proposed in \cite{zhang2018unreasonable} and the SqueezeNet \cite{squeeze} (noted $\mathcal{S}$) to evaluate the perceptual similarity between the raw processed videos $P$ and output $O$ videos.
\[
D_{perceptual}(P,O) = \frac{1}{T-1}\sum_{t=2}^T \mathcal{S}(O_t,P_t)
\]
We discuss this metric more deeply in the next section and argue that this metric fails to correctly capture localized style deviations or brightness alteration.
\subsection{Datasets and training}
For comparability and convenience, we use the dataset made available by \cite{lai2018learning} which is constituted of videos from DAVIS-2017 dataset \cite{Perazzi2016} and videvo.net videos. The height of each video in the training set is scaled to 480 while preserving the aspect ratio. The training set contains in total 25,735 frames.
The applications considered in the dataset are the following: artistic style transfer, Colorization, Image enhancement, Intrinsic image decomposition, and Image-to-image translation.

Results presented hereafter are those obtained after training with values of lambdas set empirically : $
\lambda_p=10, \lambda_{SP} = 10, \ \lambda_{st}=100, \  \lambda_{lt}=100, \ \lambda_{PP} = 100, \ \lambda_{rank} = 0.00001
$. 
We set the number of epoch to 100, each composed of 1000 batches of size 4. The sequences considered in every batch are composed of 5 successive frames each, and thus 9 once the Ping Pong sequence is constructed. We consider Adam optimizer for parameter optimization.
\subsection{Qualitative evaluation}
\begin{figure}[h!]
\centering
\begin{subfigure}{0.165\textwidth}
  \centering
   \includegraphics[scale=0.5]{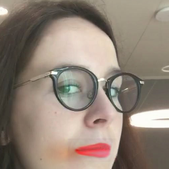}
   \caption{Lai et al.}
    \label{fig:eccv18_p13}
\end{subfigure}
\begin{subfigure}{0.165\textwidth}
  \centering
       \includegraphics[scale=0.5]{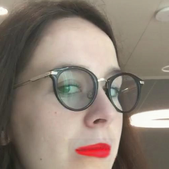}
    \caption{Ours + SP Loss}
    \label{fig:retrained_p13}
\end{subfigure}
\caption{(a) displays a frame of a video processed using make-up synthesis (lipstick) and post-processed using \cite{lai2018learning} which removes the lip flickering but at the cost of a red trail following the lips as the head moves. (b) shows the same frame post-processed using our proposed model which removes flickering without generating any trail. The rectangular shape seen around the lips in both (a) and (b) originates from the processing algorithm which we used.}
 \label{fig:retrained_vs_eccv18_trail}
\end{figure}
We propose to compare our model to the current state-of-the-art post-processing technique of \cite{lai2018learning}. The latter model successfully removes flickering artifacts in most cases but suffers from two drawbacks : (i) it tends to generate new temporal artifacts when flickering is localized on videos or when the motion of an object changes direction and (ii) the overall color of the video is often modified after using their post-processing method. An example of (i) can be seen in Fig. \ref{fig:eccv18_p13} where make-up synthesis was applied which generates flickering on the lips. Once post-processed using the model of \cite{lai2018learning} a trail following the lips appears. We believe this type of artifacts to originate from the form of the long-term loss considered in their model which forces the model to be coherent with the first frames of the sequence.

Our model visually outperforms the current state-of-the-art model in terms of style conservation and flicker removal. For instance Fig. \ref{fig:retrained_p13} shows that our model successfully removes flickering and does not generate trails as \cite{lai2018learning}. Similarly Fig. \ref{fig:cows} shows how our model manages to better remove flickering while not suffering from perceptual deterioration of the overall style of the video. For instance, in Fig \ref{fig:cows} one can see the videos post-processed using \cite{lai2018learning} becoming darker frame after frame, while our model coherently keeps the overall look of the video while not altering the brightness nor the style of the video. More examples showing the performance of our model are available in the supplementary material.
\subsection{Quantitative comparisons}
To evaluate the temporal coherence of videos we resort to the warping error metric previously discussed. Table \ref{tab:time_met} shows that our model outperforms the model of \cite{lai2018learning} for both DAVIS and videvo.net datasets for a large majority of applications when the style preserving loss is considered and for all test applications considered when not including the style preserving loss.
Let us note that we obtain better results than \cite{bonneel:hal-01264081} on applications where they outperformed \cite{lai2018learning}. Also note that the average warping error for the raw processed frame are $0.052$ for the DAVIS dataset and $0.053$ for the videvo.net dataset. 

\begin{table}[h!]
\caption{\textbf{Warping Error} of post-processing models }
    \centering
    %\footnotesize
    \tiny
       \begin{tabular}{l|c c c | c c c}
    \toprule
      \multirow{3}{*}{Task} & \multicolumn{3}{c}{DAVIS} & \multicolumn{3}{c}{VIDEVO.NET} \\
        &  \multirow{2}{*}{Lai et al.} & \multirow{2}{*}{Ours} & Ours +  & \multirow{2}{*}{Lai et al.} & \multirow{2}{*}{Ours} & Ours + \\
        & & & SP loss & & & SP loss \\
     \midrule
     \textbf{WCT/antimono}  & $0.031$ & $\mathbf{0.017}$ & $0.024$  & $0.022$ & $\mathbf{0.013}$ & $0.019$  \\
     \textbf{WCT/asheville} & $0.059$ & $\mathbf{0.029}$ & $0.043$ & $0.047$ & $\mathbf{0.024}$ & $0.037$\\
     \textbf{WCT/candy} & $0.045$ & $\mathbf{0.025}$ & $0.039$ & $0.034$ & $\mathbf{0.020}$ & $0.031$ \\
     \textbf{WCT/feathers} & $0.039$ & $\mathbf{0.022}$ & $0.036$ & $0.034$ & $\mathbf{0.021}$ & $0.034$\\
     \textbf{WCT/sketch} & $0.030$ & $\mathbf{0.015}$ & $0.024$ &  $0.030$ & $\mathbf{0.014}$ & $0.023$ \\
     \textbf{WCT/wave}  & $0.036$ & $\mathbf{0.020}$  & $0.032$ & $0.057$ & $\mathbf{0.017}$ & $0.027$ \\
     \textbf{Fast-neural-style/princess}& $0.060$ & $\mathbf{0.036}$ & $0.060$  & $0.060$ & $\mathbf{0.039}$ & $0.065$\\
     \textbf{Fast-neural-style/udnie} & $0.023$ & $\mathbf{0.012}$ & $0.017$ & $0.017$ & $\mathbf{0.009}$ & $0.013$ \\
     \textbf{DBL/expertA}  & $0.013$ & $\mathbf{0.008}$ & $0.011$ & $0.012$ & $\mathbf{0.008}$ & $0.011$ \\
     \textbf{DBL/expertB} & $0.011$ & $\mathbf{0.007}$ & $0.009$ & $0.009$ & $\mathbf{0.006}$ & $0.010$\\
     \textbf{Intrinsic/reflectance}  & $0.010$ & $\mathbf{0.006}$ & $0.009$  & $0.010$ & $\mathbf{0.007}$ & $0.010$ \\
     \textbf{Intrinsic/shading} & $0.009$ & $\mathbf{0.005}$ & $0.007$ &$0.007$ & $\mathbf{0.005}$ & $0.007$ \\
     \textbf{CycleGAN/photo2ukiyoe}  & $0.020$ & $\mathbf{0.013}$  & $0.019$ & $0.017$ & $\mathbf{0.012}$ & $0.018$ \\
     \textbf{CycleGAN/photo2vangogh} & $0.027$ & $\mathbf{0.020}$ & $0.028$ &$0.025$ & $\mathbf{0.017}$ & $0.028$ \\
     \textbf{Colorization} & $0.012$ & $\mathbf{0.007}$ & $0.009$ & $0.009$ & $\mathbf{0.006}$ & $0.008$ \\
     \textbf{Colorization} & $0.010$ & $\mathbf{0.006}$ & $0.008$ & $0.008$ & $\mathbf{0.006}$ & $0.008$ \\
     \midrule 
     \textbf{Average} & $0.027$ & $\mathbf{0.015}$ & $0.023$ & $0.023$ & $\mathbf{0.014}$ & $0.021$\\
     \bottomrule 
\end{tabular}
    \label{tab:time_met}
\end{table}

When improving temporal consistency in a video, one usually faces a trade-off between making a video more coherent and not altering the overall style of the video. 
The metric presented in section \ref{pd} fails to fully capture style conservation. For instance, one major drawback of this metric is its inability to correctly account for new temporal artifacts appearing: in Fig. \ref{fig:retrained_vs_eccv18_trail} and the video available in the supplementary material one can see that the post-processing model of \cite{lai2018learning} generates a trail following the lips, decreasing the resemblance between the original processed video and the post-processed video. However, the perceptual distance seems to favor \cite{lai2018learning} rather than our proposed model despite such phenomenon ($0.026$ for the post-processing model of \cite{lai2018learning} and $0.052$ for our proposed model). Similarly, \cite{lai2018learning} tends to alter the overall style of videos by either changing the contrast or the brightness of videos, while our model successfully avoids such pitfalls (see Fig. \ref{fig:brightness} where both post-processing models obtain a perceptual distance of $0.038$ over the whole video) but the metric can still favor the competing model. We believe that the metric only captures blurriness and major color drift, but fails to measure true resemblance and style preservation. Our model obtains an average perceptual distance over all applications of respectively $0.058$ and $0.056$ with and without our style preserving perceptual loss for the DAVIS dataset, as well as $0.051$ and $0.053$ for the videvo.net dataset, while \cite{lai2018learning} obtained respectively $0.017$ and $0.012$.

\subsection{Discussion}
\begin{figure}[h!]
    \centering
    \includegraphics[scale=0.4]{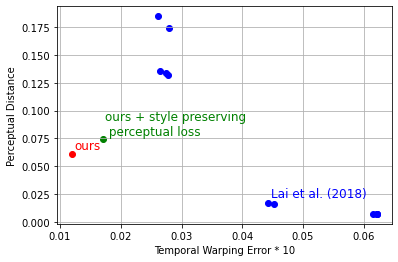}
    \caption{Both models are compared using the temporal warping error and the perceptual distance for the application fast-neural-style/udnie. Our model significantly outperforms \cite{lai2018learning} in terms of warping error and obtains slightly higher perceptual distance.}
    \label{fig:graph}
\end{figure}
\begin{figure}[h!]
    \centering
    \includegraphics[scale=0.2]{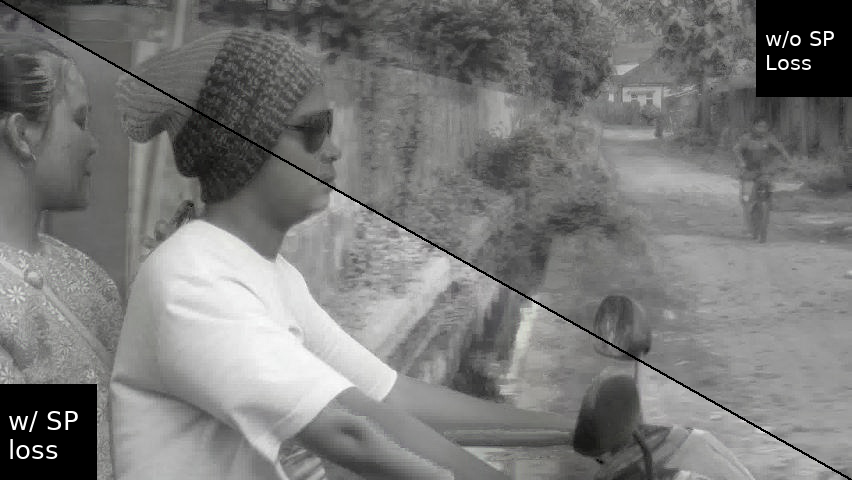}
    \caption{Without our style preserving perceptual loss the video tends to become darker when post-processed.}
    \label{fig:brightness}
\end{figure}

\noindent As seen in Fig. \ref{fig:graph}, we obtain the lowest temporal warping error at the cost of a slightly larger perceptual distance than \cite{lai2018learning}. Note that when making their hyperparameters vary, \cite{lai2018learning} manage to improve their temporal coherence at the cost of a much higher perceptual distance, but they never attain as low perceptual distance as our model. The blue dots in Fig. \ref{fig:graph} correspond to the model of \cite{lai2018learning} for several sets of hyperparameters.
When including the style preserving perceptual loss, we lightly deteriorate both the perceptual distance and temporal warping error. 
However as seen in Fig. \ref{fig:brightness} the style preserving perceptual loss reduces deterioration of the brightness throughout videos and style deviation. This supports the idea that the perceptual distance fails to fully capture style preservation. For instance, both our model and the model of \cite{lai2018learning} can alter the brightness of videos and our proposed loss manages to reduce such issue.
\section{Conclusion}
\label{conclusion}
We discussed questions regarding temporal consistency linked to independently processing videos frame by frame using image trained algorithms. We put forward a new model to remove flickering artifacts, it includes recent techniques proposed in approaches orthogonal to our work in the field of video processing, as well as a novel loss to circumvent problems linked to style deterioration throughout videos.
Thanks to these modifications, we obtain state-of-the-art results regarding flickering removal and also improve the visual output of the post-processed videos.
We believe that future work in the field of video temporal post-processing ought to focus on algorithms that do not necessitate to have the original non-flickering video for inference.


\bibliographystyle{IEEEbib}
\bibliography{refs.bib}

\end{document}